\documentclass[letterpaper]{article} 
\usepackage{aaai24}  
\usepackage{times}  
\usepackage{helvet}  
\usepackage{courier}  
\usepackage[hyphens]{url}  
\usepackage{graphicx} 
\def\UrlFont{\rm}  
\usepackage{natbib}  
\usepackage{caption} 
\usepackage{algorithm}
\usepackage{algorithmic}
\usepackage{enumerate}
\usepackage{enumitem}
\usepackage{booktabs, multirow} 
\usepackage{soul}
\usepackage[table]{xcolor} 
\usepackage{changepage,threeparttable} 
\usepackage{amsmath}
\newcommand\norm[1]{\left\lVert#1\right\rVert}
\usepackage{newfloat}
\usepackage{listings}
\DeclareCaptionStyle{ruled}{labelfont=normalfont,labelsep=colon,strut=off} 
\floatstyle{ruled}
\newfloat{listing}{tb}{lst}{}
\floatname{listing}{Listing}
\title{Mimicking the Maestro: Exploring the Efficacy of a Virtual AI Teacher in Fine Motor Skill Acquisition}
\author {
    Hadar Mulian\textsuperscript{\rm 1},
    Segev Shlomov\textsuperscript{\rm 1},
    Lior Limonad\textsuperscript{\rm 1},
    Alessia Noccaro\textsuperscript{\rm 2},
    Silvia Buscaglione\textsuperscript{\rm 3}
}
\affiliations {
    \textsuperscript{\rm 1}IBM Research - Israel\\
    \textsuperscript{\rm 2}Neurorobotics Lab, School of Engineering, Newcastle University, Newcastle Upon Tyne, United Kingdom\\
    \textsuperscript{\rm 3}NEXT Lab, Universita’ Campus Bio-Medico di Roma, Rome, Italy\\
    Hadar.Mulian@ibm.com, Segev.Shlomov1@ibm.com, 
    liorli@il.ibm.com, alessia.noccaro@newcastle.ac.uk, s.buscaglione@unicampus.it
}

\usepackage{bibentry}
\usepackage{xcolor}
\usepackage{amsmath,amssymb}

\begin{document}

\maketitle
\newif\ifshowcomments
\showcommentstrue
\ifshowcomments
\newcommand{\mynote}[2]{\fbox{\bfseries\sffamily{#1}}
 {\small$\blacktriangleright$\textsf{#2}$\blacktriangleleft$}}
\else
\newcommand{\mynote}[2]{}
\fi
\newcommand{\hadar}[1]{\textcolor{blue}{\mynote{Hadar}{#1}}}
\newcommand{\lior}[1]{\textcolor{purple}{\mynote{Lior}{#1}}}
\newcommand{\segev}[1]{\textcolor{purple}{\mynote{Segev}{#1}}}
\newcommand{\alessia}[1]{\textcolor{red}{\mynote{Alessia}{#1}}}
\newcommand{\silvia}[1]{\textcolor{green}{\mynote{Silvia}{#1}}}
\begin{abstract}
    Motor skills, especially fine motor skills like handwriting, play an essential role in academic pursuits and everyday life. Traditional methods to teach these skills, although effective, can be time-consuming and inconsistent. With the rise of advanced technologies like robotics and artificial intelligence, there is increasing interest in automating such teaching processes. In this study, we examine the potential of a virtual AI teacher in emulating the techniques of human educators for motor skill acquisition. We introduce an AI teacher model that captures the distinct characteristics of human instructors. Using a reinforcement learning environment tailored to mimic teacher-learner interactions, we tested our AI model against four guiding hypotheses, emphasizing improved learner performance, enhanced rate of skill acquisition, and reduced variability in learning outcomes. Our findings, validated on synthetic learners, revealed significant improvements across all tested hypotheses. Notably, our model showcased robustness across different learners and settings and demonstrated adaptability to handwriting. This research underscores the potential of integrating Imitation and Reinforcement Learning models with robotics in revolutionizing the teaching of critical motor skills.
\end{abstract}

\section{Introduction}

Fine motor skills, such as handwriting, are pivotal not only in the academic journey of an individual but also in daily life tasks in developing core capabilities such as physical coordination, rhythm, stamina, and posture~\cite{handwriting1992,scordella2015role,swain2018improving}. The ability to effectively teach and improve these skills has long been the domain of human educators, who rely heavily on personalized feedback and repetitive exercises. Traditional teaching methods, although proven effective, can take a long time and sometimes lack the consistency needed to master a skill~\cite{gul2014efficacy, irby2004time}.

In recent years, advancements in robotics and artificial intelligence have spurred interest in automating some teaching processes, leveraging these technologies and their interplay with human interactions. Automation, especially through robots~\cite{Social-robots2018}, offers consistent, repetitive, high-frequency feedback~\cite{krebs1998robotaided}, which might be challenging for a human instructor, especially in larger classroom settings or for the teaching of competencies such as motor skills that require extensive repetition for mastery.

Our research examines whether a virtual AI teacher can emulate real educators and help learners master motor skills. We developed an AI teacher trained on a fundamental follow-the-cursor task, mimicking human teaching.
Motivated by the way learners such as kids learn fine motor skills, we tested our model using RL-based students, on different motor tasks. We then implemented an environment that captures the teacher-learner interactions and enables us to test whether our model improves the performance of learners. We have established four guiding hypotheses, which we anticipate every virtual teacher to uphold; Firstly, a learner's performance improves after each instructional session with a virtual teacher, in contrast to learners who pursue learning without assistance. Additionally, over the span of instructional sessions, learners guided by a virtual instructor consistently outperform those who embark on independent learning. Third, when learners are paired with a virtual teacher, they grasp motor skills at an appreciably faster rate. Lastly, the variability in learner outcomes diminishes when education is facilitated by the virtual teacher, and this fluctuation continues to contract as the learning advances.

We tested and evaluated these hypotheses on synthetic learners and showed significant results on all four of them. Moreover, our results demonstrate model robustness across different learners, among various environment settings, and generalizability to handwriting as a more concrete motor skill.

We note that in the broader context, AI has been at the forefront of integrating technology into education, continually seeking methods to make learning more effective and personalized~\cite{Woolf2008,mukherjee2013learner}. As a part of this overarching mission, our research falls under a broader initiative that aims at leveraging robotics in the realm of education. To address our research question and test the proposed hypotheses, we have embarked on training an Imitation Learning (IL) based motion-control model to function as a virtual instructor for motor skill acquisition.

To conclude, our main contribution:
\begin{itemize}[leftmargin=*]
\item
To the best of our knowledge, we are the first to develop an AI teacher for fine motor learning tasks that captures the 
characteristics of real human teachers.
Our model can increase the performance of learners, shorten their learning time, and lower the overall performance variance.
\item We implemented an adaptable environment for training and assessing a GAIL model to teach a motor skill, and used it to facilitate the training of a virtual teacher, equipping it with the ability to train on a series of foundational shape-creation skills. 
Subsequently, the environment serves as a testbed platform to evaluate the model's effectiveness in teaching a motor skill. It employs RL to simulate a series of teacher-learner interactions.
\item We open source both the model and the environment to the community. Both are available on 
\url{https://github.com/IBM/SAX/tree/main/Conbots/EAAI24}
\end{itemize}

Through the course of this paper, we explore the methodologies employed, the challenges encountered, and the results derived from this innovative approach. We aspire to provide insights on whether robotics, powered by IL models trained to mimic human behavior, can truly revolutionize the learning of specific motor skills such as handwriting, playing music, driving, and alike.

\section{Related Work}

Developing a virtual agent to teach motor skills is an interdisciplinary topic that resides at the intersection of AI, human-computer interaction and robotics. Using robots 
to teach motor skills very often finds its roots in rehabilitation robotics, especially for patients recovering from events like strokes~\cite{Gupta2006,Mataric2007-um,Lambercy2007-zz}. More recently, there has been an increasing interest in using robots for general education and training outside the realm of rehabilitation~\cite{Alemi2015-qt,Chernova2014,Mubin2013, Rozo2013}. However, existing work is limited in the context of data-driven applications, for the automation of teaching of motor skills.
The predominant research focus in related areas, as also presented in~\cite{caramiaux2020machine}, has been on enabling models to acquire motor skills themselves, or to stimulate ongoing feedback upon learner's actions~\cite{bonneton2020can, ecalle2021spatial}. However, none of the previous work is focused on the elicitation of the teacher's behavior to be replayed upon teaching these skills to others.

The scarce set of studies that focus on augmented human-robot interaction, describe a series of similar experiments.
\cite{takagi2018haptic} introduces a reactive robot partner that interacts with a learner. Together they share movement goals via the forces they perceive. 
The coupling dynamics, represented by the stiffness of the virtual elastic band (hard, medium, or soft), determine the capacity of communicable information.
This approach is later shown in \cite{ivanova2020motion} as a favorable motion assistance paradigm in tracking tasks, when compared to passive trajectory guidance, or a human partner, and \cite{ivanova2022interaction} which also supports a similar approach.

Our work employs Generative Adversarial Imitation Learning (GAIL), which has been recently popularized as an effective method for acquiring motor gestures from expert demonstration~\cite{ho2016generative}. It leverages the power of Generative Adversarial Networks (GANs) to train an agent to imitate an expert behavior on a given task; Using two neural networks, the training process contains two repeated steps: (1) a generator policy generates a trajectory, attempting to mimic the expert actions, and (2) a discriminator that attempts to distinguish between the expert trajectory and the one generated by the agent, encouraging the agent to improve its imitation skills over time. 



To the best of our knowledge, there is little to no work on developing models for teaching robots how to teach humans motor skills. Furthermore, none of the previous work has been focused on leveraging contemporary imitation learning methods such as GAIL to enable the realization of a digital apprentice that can facilitate and ultimately automate the teacher-learner interaction for the acquisition of motor skills, to the extent that it can be employed to fully replace the role of a human teacher. Further to this, our work also adds a second tier of novelty in developing RL-based simulated student learners to boost the assessment of the effectiveness of the developed teacher model, 
alleviating the time consuming and tedious process of empirical assessment.

\section{The Model}




Central to our research is the design and implementation of an advanced learning model explicitly tailored for motor skill acquisition. Our approach has three main components, with each designed to optimize the effectiveness of the system. We next outline the core elements of our methodology.

The essence of our model's learning capabilities is in Imitation Learning techniques of real teacher-learner interactions. By leveraging the principles of adversarial networks, our approach sets one network (the imitator) to endure replicating expert motor-skill demonstrations, while its counterpart (the discriminator) evaluates the imitated output against the recorded human expert demonstrations. This iterative adversarial interplay refines the imitation capabilities over time, resulting in motor-skill trajectories that closely mirror those demonstrated by human educators. We note that this GAIL approach sidesteps the task of manually crafting reward functions. Instead, by harnessing real-world demonstrations, our model adeptly captures the subtleties and nuances inherent in the behavior of a real teacher.

To ascertain the viability and effectiveness of the AI teacher, we introduced an RL-based method for emulating real learners. Given the inherent challenges and impracticality tied to extensive real-world testing, we engineered virtual learners. These digital learners, established by reinforcement learning principles, emulate a diverse array of learner profiles. Their behaviors and learning trajectories, drawn from a mix of novice and expert learners, provide a treasure trove of simulation data. Subjecting these virtual learners to both traditional and our novel teaching methods afforded us a comparative lens, ensuring that our findings are robust.

To do so, we crafted a Learner-Teacher interaction Environment. Recognizing that the dynamics between a learner and teacher are multifaceted, our environment was meticulously designed to simulate the real-world challenges and nuances of this relationship. This custom-built space allowed for the execution of motor tasks under varying conditions, mirroring the complexities learners might face in real-world scenarios. At the core of this environment is an interactive feedback loop, enabling the virtual teacher and the virtual learners to perform tasks and receive immediate feedback. This real-time feedback system, paired with the responsive nature of the GAIL model, augments the overall learning process, driving iterative improvement. 


Together, these integrated components fashion a model and environment that seeks to not just emulate but enhance the traditional teaching paradigm.

\subsection{Hypotheses} \label{our_goal}
Our main research question stands for whether a virtual teacher can learn how to mimic a real teacher in helping learners acquire motor skills.
This question has manifested itself in the following four hypotheses:

\begin{enumerate}[label=H\arabic*:, leftmargin=*]
  \item The performance of a learner guided by a virtual teacher is better following each learning session, compared to a learner who is not guided by a virtual teacher.
  \item The overall performance of a learner guided by a virtual teacher, across a series of learning sessions, is better than a learner who is not guided by a virtual teacher.
  \item The learner acquires a motor skill significantly faster when it is guided by a virtual teacher compared to achieving the same level of competence not being guided by a virtual teacher. 
  \item Performance variance is lower when a learner is guided by a virtual teacher compared to a learner who is not guided by a virtual teacher, and further decreases as learning progresses.
\end{enumerate}

To address our research question and test the hypotheses, we first recorded a series of ``Follow-The-Cursor'' (FC) teacher-learner sessions. Using this data, we applied an imitation-learning method to develop a GAIL model, designed to act as a virtual teacher for motor skill acquisition. Subsequently, we established an RL-based environment to simulate a virtual population of learners. Using this environment, we systematically compared the performance of learners assisted by our developed model against those who learned independently. This comparison was centered on each of the hypotheses listed above.

\section{Experiments and Settings}

\begin{figure}[ht]
    \centering
    \includegraphics[width=0.95\columnwidth]{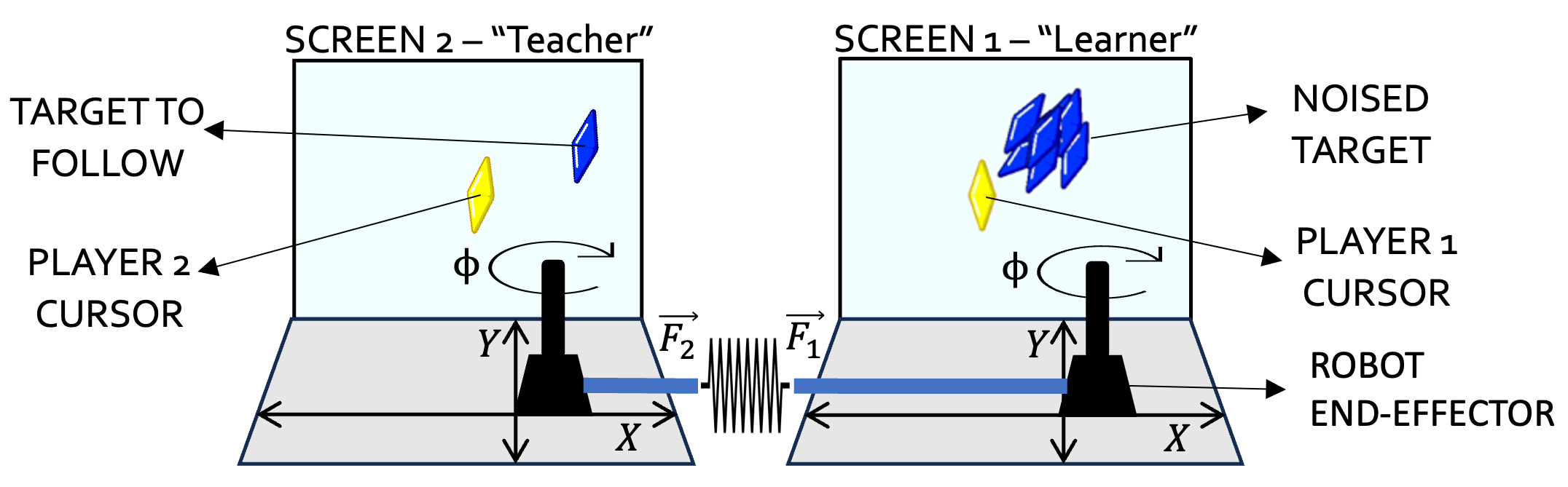}
    \caption{Follow-the-cursor experiment}
    \label{fig:FC-experiment}
\end{figure}

To drive model development we used data collected in a previous work~\cite{noccaro2024} at NEXTLab (Universita' Campus Bio-medico di Roma). In that study, 20 healthy subjects (aged 22.5 ± 2.6 years) volunteered to participate in a follow-the-cursor (FC) experiment, after having signed a written informed consent. The subjects were grouped into pairs, instructing each to perform a task twice. In the task, subjects were asked to follow a target cursor that was presented to them on a screen, controlling their own cursor by moving mutually coupled robot end-effectors. The target was to be followed on three coordinates: $x, y, \phi$ as illustrated in Figure~\ref{fig:FC-experiment}.

The physical connectivity between the two end-effectors was defined by an elastic force based on their relative positions and connection stiffness, as described by the equations below. The positions of the learner-teacher pair at any time are represented by $\Lambda_t$ and $\Upsilon_t$ respectively:
\begin{align*}
    F_\Lambda & = K_\Lambda ( \Upsilon_t - \Lambda_t ) + D \cdot \Lambda_t\\
    F_\Upsilon & = K_\Upsilon ( \Lambda_t - \Upsilon_t ) + D \cdot \Upsilon_t
\end{align*}
where 
 $F_\Lambda$ and $F_\Upsilon$ are the forces on the end-effectors
, $K_\Lambda$ and $K_\Upsilon$ are stiffness constants for the end-effectors
, $\Lambda_t$ and $\Upsilon_t$ represent the positions of the two end-effectors or cursors.
, and $D$ is a damping constant, applied linearly to the position.

The force applied on each end-effector is proportional to the difference in position between the two end-effectors, multiplied by the stiffness constant $K$. This yields a spring-like behavior, where the force increases as the distance between the two positions increases, trying to bring them closer. Damping acts to reduce the velocity of a system, mitigating the resistance based on the position of the end-effectors. In the experimental session, stiffness levels were manipulated at two levels for each subject, high (H) and low (L), where all four combinations HH, LL, HL, and LH were tested with linear stiffness $(x,y)$ and rotational stiffness $(\phi)$ values as listed in Table~\ref{tab:stiffness}. 

\begin{table}[!htp]\centering
\setlength{\tabcolsep}{4pt}
\small
\begin{tabular}{l|cc|cc}\toprule
Modality & $K_\Lambda^x=K_\Lambda^y$ & $K_\Upsilon^y=K_\Upsilon^x$ & $K_\Lambda^\phi$ & $K_\Upsilon^\phi$ \\
 & [N/m] & [N/m] & $[Nm/rad]$ & $[Nm/rad]$ \\\midrule
HH & 180 & 180 & 9 & 9 \\
LL & 60 & 60 & 6 & 6 \\
LH & 60 & 180 & 6 & 9 \\
HL & 180 & 60 & 9 & 6 \\
\bottomrule
\end{tabular}
\caption{Stiffness levels}\label{tab:stiffness}
\end{table}


To induce skill differences between partners, one subject's target cursor was surrounded by noisy cursors deviating from the real target using a Gaussian distribution. In the next session, roles switched. The partner without noise was deemed the ``teacher'', and the one with noise, the ``learner''.

Before each of the two experimental sessions, participants went through familiarization and baseline phases. In the familiarization phase, subjects undertook six solo tasks; half had visual noise and half did not, presented in a random sequence. The baseline phase had 20 solo trials, setting a performance standard for each participant under the same visual noise condition they would later encounter in the experimental phase. In the experimental sessions, there were 80 coupled trials, 20 for each stiffness setting. Every trial consisted of a 30-second interaction and a 10-second rest. Data included details of the cursor position, orientation, and exerted force for both participants, along with stiffness and modality of connectivity.

These experimental sessions were recorded and adjusted as a $250$ step-long trajectory, to be later fed by the GAIL model as expert behavior recordings, used during training by the discriminator.

Given the challenge of modeling $3D$ actions on a plane, the recordings captured only the x and y axes data. 



\subsection{Preliminaries}

Let us denote the $2D$ locations of the target, teacher, and learner with $\Psi,\Upsilon,\Lambda$ respectively. The location on a specific axis is denoted by square brackets with the axis specification, i.e., $\Lambda[0]$ marks the location of the learner on the x-axis, and $\Lambda[1]$ marks its location on the y-axis.
\begin{itemize}[leftmargin=*]
    \item $\mathbf{\square^{+}}$: (superscript) learner and teacher are connected while learning. Notation is added to both locations.
    \item $\mathbf{\square^{-}}$: (superscript) learner and teacher are not connected while learning. Notation is added to both locations.
    \item $\mathbf{a_t}$: an action on the board, associated with the movement of either the teacher or learner, stated in a superscript. This way, $a_t^\Lambda$ marks an action of the learner at step $t$.
\end{itemize}

\subsection{Environment Setup}

An essential component for training GAIL models is a simulated environment 
mirroring the conditions of the environment in which expert data was originally recorded, as illustrated in Figure \ref{fig:FC-experiment}. 
This imperative arises from the fundamental underpinning of GAIL, wherein the learner's capacity to pursue a predefined target within the simulated environment,
facilitates iterative experimentation and precision refinement of its actions. Ultimately, this process aims to yield trajectories that exhibit a high degree of similarity and are nearly indistinguishable from the expert-generated trajectories, as discerned by the discriminator. 
Effectively, we built a virtual environment as a continuous $60\times60$ board. The environment allowed two-dimensional movements of the participants up to a padding of $3$ units around its edges. 
To recreate similar tracking settings to the ones shown during the experiment, for training purposes, the target of the simulated environment was updated dynamically for each time step, taking the form of sines functions and with movements adjusted to be contained within the board borders (exact details can be found in the appendix).

We define a learning step by a single model update in response to a single environment update, which includes both the target and the teacher/learner agent. An episode is characterized by 250 consecutive target updates.



To replicate the skill level difference between the teacher and the learner as was also done in the recording sessions, we obscured the target cursor with dummy targets; while the teacher has visibility of the true locations, $\Psi_t,\Upsilon_t,\Lambda_t$, the learner sees $\tilde{\Psi}_t,\Upsilon_t,\Lambda_t$, where $\tilde{\Psi}_t$ is an obscured version of the true target, drawn from one of two distributions: 
1) \textbf{Normal noise}: The noised target was drawn from a normal distribution $\tilde{\Psi}_t\sim \mathcal{N}(\Psi_t,\,\sigma^{2})$, centered at the true location of the target. 
2) \textbf{Uniform noise}: The corrupted target was uniformly chosen from the contour of a circle with radius $r$, centered at the true location of $\Psi_t$. 

\textbf{Location Updates:} The locations of the teacher and learner agents were updated at each time step according to their respective model predictions, and the counter-applied forces in case they are connected. The teacher according to the developed model and the learner according to the reinforcement learning policy.
Forces:
\begin{align*}
    F_\Lambda & = K_\Lambda \cdot (\Upsilon_t - \Lambda_t) \\
    F_\Upsilon & = K_\Upsilon \cdot (\Lambda_t^+ - \Upsilon_t^+)
\end{align*}

Finally, location update at time $t+1$ takes one of the two forms - 
\begin{itemize}
    \item Teacher and Learner are connected:
    \begin{align} \label{s_loc_update_conn}
        \Lambda_{t+1}^{+} & = \Lambda_{t+1}^{+} + (F_\Lambda + a_t^\Lambda) \cdot c 
        \\ \label{t_loc_update_conn}
        \Upsilon_{t+1}^{+} & = \Upsilon_{t}^{+} + F_\Upsilon + a_t^\Upsilon 
    \end{align}
    
    \item Teacher and learner are not connected:
    \begin{align} \label{s_loc_update_not_conn}
        \Lambda_{t+1}^{-} & = \Lambda_{t+1}^{-} + a_t^\Lambda \cdot c 
        \\ \label{t_loc_update_not_conn}
        \Upsilon_{t+1}^{-} & = \Upsilon_{t}^{-} + a_t^\Upsilon 
    \end{align}
    
\end{itemize}

Where $c$ is a mitigation factor, accounting for 
the differences in writing speeds between the teacher and learner.

\subsection{Agent Models}

\begin{figure*}[ht!]
    \centering
    \includegraphics[width=\textwidth]{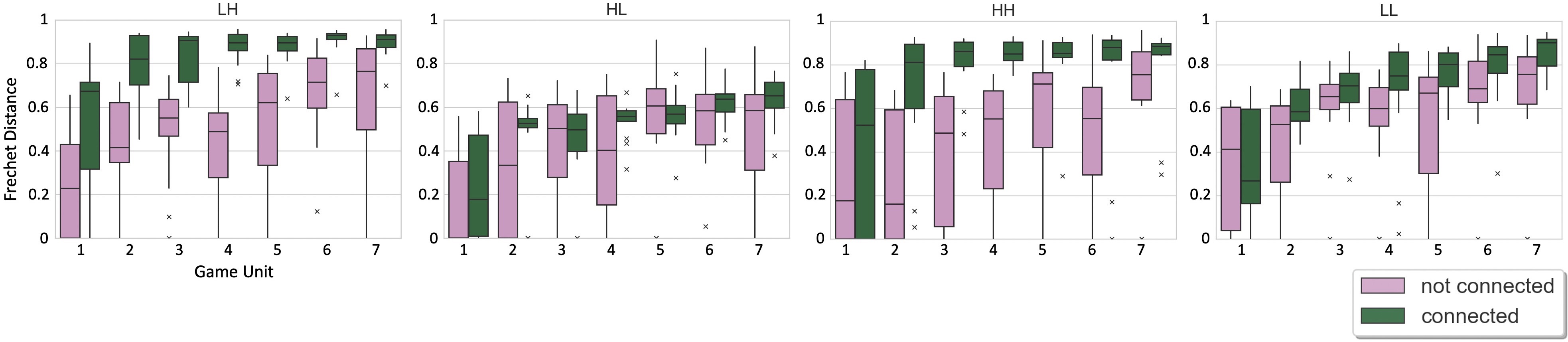}
    \caption{Learner learning curves on FC task, when training with (green) or without (pink) an attached teacher, across four connectivity modalities; From left to right: low-high, high-low, high-high, low-low. The X-axis represents the learning iteration (game unit).}
    \label{fig:boxplots_all_shapes}
\end{figure*}

\textbf{The Teacher Model} requires two types of data for training;  Experiment recorded data, containing the recorded teacher-learner sessions described above, and episodes to train on, which were generated according to the target location updates for the FC task, 
as in eq.\ref{FC_loc_update_gen_0} and eq.\ref{FC_loc_update_gen_1}.
        
To simulate an optimal environment for the GAIL model to learn a teacher model, one has to recreate similar settings to the real experiment, from which the experiment data was recorded. Therefore, to account for the teacher-learner configuration of the environment, we used a learner heuristic, present in the environment during the teacher model training. 

The learner strategy was straightforward and computed for each time step, $t$, as the following:
$a_t^\Lambda= \frac{\tilde{\Psi}_t - \Lambda_t}{\norm{ \tilde{\Psi}_t-\Lambda_t }_2} \cdot f$
where the direction of movement is the unit vector of the distance between the noisy target and learner, and the force of action $f$, is chosen randomly and equals $1$ or $2$.
Final locations were updated at the same manner as equations \eqref{s_loc_update_conn} and \eqref{s_loc_update_not_conn}, with respect to the desired connectivity setup.\\
    
Ten versions of the model were created for each virtual teacher, capturing snapshots of the teacher at various stages of the training process.

    
%
    

\textbf{Simulating a Learner Agent:} 
To test our teacher, we need to evaluate it using learners. 
To sidestep the complexities and costs associated with human student experiments, which are unscalable, we opted to test against RL-based learners. We simulated learning sessions of the teacher model with a virtual learner, employing a Synchronous Advantage Actor-Critic (A2C) agent~\cite{mnih2016asynchronous}. We note that the A2C learner is considered proficient; thus, enhancing it suggests that the teacher is genuinely effective.

The learner's training sessions were conducted in an environment similar to the one aforementioned. The reward was formulated by minimizing the learner's distance to the noised target and the magnitude of force counter applied by the teacher: 
$d_t=\norm{\tilde{\Psi}_t-\Lambda}_2 + \delta \cdot \norm{F_\Lambda}_2 $, where $\delta=50$ is a scaling constant added for even weighting of the two objectives. A booster, $g=15$ was added to the reward when the learner was close enough to the noised target. Hence, the reward $r_t=\frac{1}{d_t} + g $ was returned at each time step.

\subsection{Evaluation Test Design}
To quantitatively assess the model's teaching abilities and to capture the learning curve of a learner, both with and without a teacher, model training and evaluation were alternated, repeating the following steps: 
\begin{enumerate}
    \item Train the learner A2C model with/without an attached teacher, for $1500$ learning steps, on the FC task.
    \item Evaluate the model performance on one of two motor tasks: FC and handwriting.
\end{enumerate}

The evaluation results from a single iteration of the first step will subsequently be referred to as a "game unit".

Considering the FC task as a fundamental motor activity that reflects general motor ability across various specific tasks, such as handwriting, playing music, drawing, etc., we introduced a second task to demonstrate the model's capacity to generalize. Concretely, we employed two test sets: (1) a set of arbitrarily generated FC trajectories, and (2) English letters trajectories.

To assess the model's ability to generalize its application to other, more specific motor skills, we applied the above-designed experiment to a handwriting task, using data created in-house.
The task was defined as writing letter-long episodes and was titled Write English Script Letters (WESL). Each such letter episode was created as a list of discrete locations on a grid with the environment size, ordered as a step-by-step instruction of how it should be written.
A total of $26$ such episodes were created, containing one episode per script letter. Those had an average length of 71.96 data points in comparison to the 250 long episodes for the FC task.\\
The environment was adjusted accordingly when the model was evaluated on the WESL task. Specifically, the target location was selected from the generated letters data, respective to the current letter episode and time step, $t$.

Given two trajectories, one containing the desired shape or letter trajectory, and another containing the model predictions, either made by the teacher or by the learner, we evaluated the performance using two stages, (1) Trajectory calibration - employed to neutralize aspects such as the shape overall scale and inclination of writing, in the case of WESL task, and (2) Similarity measurement - applied to the calibrated trajectories.\\
The trajectory alignment was performed using Procrustes Analysis \cite{gower1975generalized}. This method finds the optimal transformations to match the shape and orientation of two curves by using standardization, scaling, and rotation techniques. \\
As a similarity measurement, we employed Fr\`{e}chet distance \cite{frechet1957distance} on the aligned trajectories. This distance measurement offers a robust way to quantify the similarity between two curves, considering their spatial arrangement and shape characteristics. Fr\`{e}chet distance yields a value between $0$ and $1$ where $1$ marks a perfect match.

\subsection{Results}

\begin{figure*}[ht!]
    \centering
    \includegraphics[width=1\textwidth]{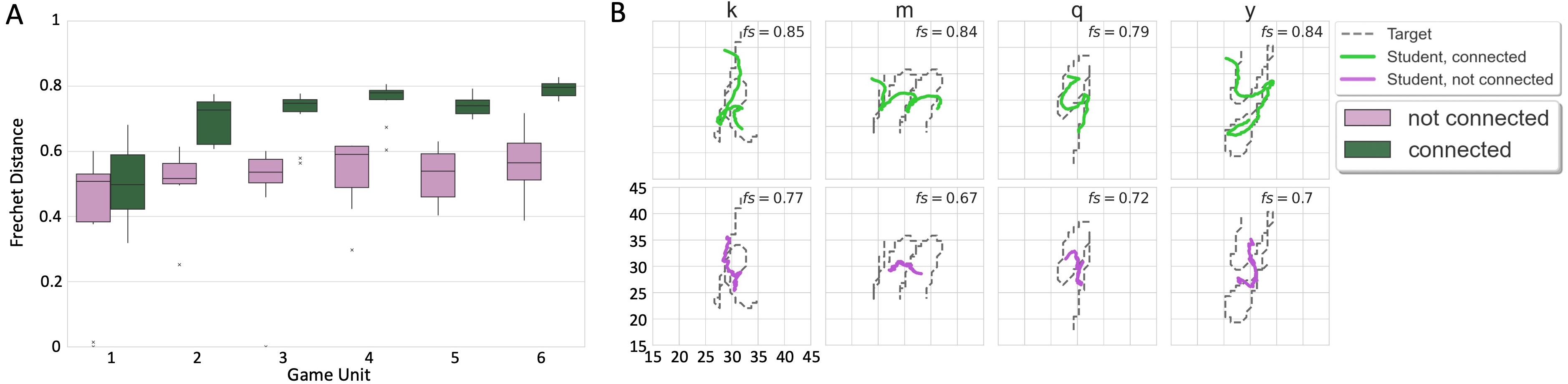}
    \caption{Learner performance on WESL task. \textbf{(A)} learning curve on WESL task, when training with (green) or without (pink) an attached teacher, trained with low-high connectivity modality. \textbf{(B)} Contour of selected letters, drawn by a randomly chosen learner when trained with (upper row figures) and without (lower row figures) a virtual teacher. Fr\`{e}chet distance computed between the target and the learner prediction is marked at the upper right of each figure. Legends for both are on the right.}
    \label{fig:letter_all_results}
\end{figure*}

For the FC skill assessment, evaluation was carried out with the aforementioned setup and was independently repeated $10$ times to demonstrate variability among learners' abilities. Figure~\ref{fig:boxplots_all_shapes} shows the results for two settings, (1) when a learner is connected to a teacher, 
(2) when the learner is not guided by a virtual teacher. Those were performed for each connectivity modality separately. While there is a noticeable difference in performance between learners with and without connectivity in each modality, our subsequent tests specifically focus on 
the LH modality, due to previous studies, such as \cite{takagi2018haptic}, suggesting it is a favorable modality for learning with the teacher's guidance. 

Results from the evaluation with the WESL task, with low-high modality, demonstrate a difference in similarity in favor of the connected settings, and overall a smaller variance in performance across all game units, as may be seen in Figure~\ref{fig:letter_all_results}.A. To get a sense of how learner agents perform at the end of the entire training session, 
we randomly picked two learners and depicted their handwriting trajectories for four letters from the WESL task. The upper row in Figure~\ref{fig:letter_all_results}.B shows the results for a learner who was trained in a connected mode, and the lower row shows the letters created by a learner who was self-trained. 
While the connected learner produced silhouettes closely resembling the target letters, the unconnected learner's performance was notably poorer, resulting in curves that lacked both range and precision. 
These contours, exemplify the difference in capability between the two learning paradigms and stress the benefits of acquiring a motor skill with the aid of a teacher.
It is also apparent that with the aid of the virtual teacher, the time needed to acquire a satisfactory level of competence is shorter. We establish the statistical significance of these results with respect to the hypotheses in the next section.



\textbf{Significance Tests} 
Each of the hypotheses presented demanded corresponding data preprocessing. Subsequently, a t-test was conducted on the Frechet distance values of the learner when connected to a virtual teacher, with low-high modality, versus the ones in the non-connected settings. The preprocessing and results for the two tasks are reported for each hypothesis as follows.

%
%
\textbf{H1}: For each game unit, a series of t-tests were conducted between the connected values versus the not-connected values after the completion of each game unit. Results for the FC task showed that the performance of a learner guided by a virtual teacher was significantly higher than the performance values of the non-guided learner, consistent across all game units (i.e., all p-values$<$0.05, see Table~\ref{tab:FC_hypth_1} in appendix). Results for the WESL task showed similar significant differences for the second game unit onwards (i.e., all p-values$<$0.05, see Table~\ref{tab:wesl_hypth_1} in appendix).\\ 
\textbf{H2}: For each repetition (a.k.a learner version), the area under the curve (AUC) of the Frechet distances across all game units was computed. The result for the FC task showed that the overall performance of the guided learner as reflected by the corresponding AUC value was significantly higher than the performance of the non-guided learner ($t=6.58$, $\mu_{AUC}^{+}=4.971$,$\mu_{AUC}^{-}=2.975$,$p<0.5\cdot10^{-5}$). The result was also consistent for the WESL task ($t=7.31$,$\mu_{AUC}^{+}=3.550$,$\mu_{AUC}^{-}=2.544$,$p<0.5\cdot10^{-5}$).\\
\textbf{H3}: Let us mark $\theta_{FC}=0.9$ as a threshold for high Frechet distance value for the FC task,  $\theta_{WESL}=0.75$ as a threshold for the WESL task. For each experiment repetition, let us mark the first game unit where the distance was higher than $\theta$ as the population values. Respective to the $\theta$ value, the result for the FC task showed that the motor skill was acquired significantly faster by the guided learner compared to the number of game units the motor skill was acquired by the non-guided learner ($t=6.20$,$\mu_{\theta}^{+}=4.307$,$\mu_{\theta}^{-}=9.692$,$p<0.5\cdot10^{-5}$). The same result was also consistent for the WESL task ($t=15.84$,$\mu_{\theta}^{+}=3.3$,$\mu_{\theta}^{-}=10.0$,$p<0.5\cdot10^{-10}$).\\
\textbf{H4}: We mark learning convergence at game unit 4 for the FC task and at game unit 3 for the WESL task. For each of the remaining game units, we computed the standard deviation as a population value. Respectively, the result for the FC task showed a lower average in the standard deviation for the guided learner compared to the non-guided learner ($t=7.38$,$\mu_{SD}^{+}=0.07$,$\mu_{SD}^{-}=0.27$,$p<0.001$). The result was also consistent here for the WESL task ($t=3.35$,$\mu_{SD}^{+}=0.036$,$\mu_{SD}^{-}=0.086$,$p<0.05$).
\subsection{Model Robustness}
In our pursuit of a well-rounded and effective virtual teacher, understanding the model's robustness under varying conditions is paramount. The value of a truly robust model is its ability to adapt, perform, and maintain efficacy under a variety of circumstances. To examine our virtual teacher robustness, we conducted extensive robustness tests, under diverse settings, to emulate different real-world scenarios. We introduced learners with different initial skills, learning rates, and attention (noise) levels, and tested each virtual teacher on them across the two tasks.

\textbf{Settings}:
Tasks: FC and handwriting.
Noise Values: Gaussian and random.  We injected the noise into the learners' actions to emulate occasional errors or lack of focus. Multiple noise variance values were tested, simulating everything from a near-perfect learner to one with significant challenges.
Iterations (game unit): All $7$ game units were tested. The learner sessions ranged from short to prolonged, reflecting differences in stamina and training duration.

Overall we tested 64 different settings and performed 32 comparisons. Each such comparison consisted of 10 distinct learners with the same settings.

\textbf{Results}:
The complete robustness results of all settings for both tasks can be found in the appendix in Tables~\ref{tab:wesl_robust} and~\ref{tab:FC_robust}. The average improvement (i.e., the difference between the Frechet distance) between connected and not-connected learners across all settings are $25.8\%$ and $15.3\%$ for the FC task and the WESL task respectively, where the standard deviations are $0.05$ and $0.025$. The maximal difference between every two comparisons is $39\%$ (FC) and 20.1\% (WESL), and the minimal improvement is $17\%$ and $11\%$ respectively. 

The results consistently demonstrated that learners connected to our AI teacher showed better performance and faster skill acquisition, irrespective of their starting conditions. While learners with higher noise values and shorter sessions generally took longer, the assistance of the virtual teacher consistently outperformed the autonomous learning approach in the two motor learning tasks.

\section{Applications in Education}
Our model was trained on a fairly simple, yet fundamental, task of following a cursor across a board. It demonstrated versatile teaching abilities in acquiring motor skills, allowing us to firmly believe the virtual model can be adapted to teach other motor skills, particularly those involving gentle motion and precision, such as writing of numbers, or geometric shapes.
Another useful application lies in the Art domain. That is, motor tasks also encompass drawing, painting, and design. Beginners in art or design courses often grapple with basic techniques like creating consistent line weights, mastering brush strokes, or executing precise cuts in craft work. An AI teacher model can break down these techniques into manageable tasks, offering learners real-time feedback as they practice. This immediate feedback loop can accelerate skill acquisition, boost confidence, and allow art educators to focus on teaching creativity and higher-order thinking rather than basic mechanics.

The idea can also be adapted, in the music domain. Learning to play a musical instrument requires mastering intricate motor skills, e.g. finger placement on a violin, breath control for a trumpet or hand coordination on drums. An AI teacher model can boost the learning process providing learners with feedback on their technique. For beginners, the teacher could guide finger placements or bowing techniques. For advanced learners, it might offer feedback on the nuances of their technique. This application can democratize access to quality music instruction, especially when human instructors might be scarce or expensive.

In addition, we argue that an AI teacher model can also be used in Virtual Reality (VR) training. As these technologies gain traction in educational settings, there's a growing need for users to interact with virtual objects using hand gestures, controllers, or styluses. An AI teacher model can be embedded into these virtual environments to teach users the most effective ways to perform motor tasks. For instance, in a VR-based art class, the AI can instruct on sculpting techniques or in an AR-based assembly training, guide on proper hand movements for efficient assembly. This melding of VR/AR with AI-based motor task training can make virtual learning environments more intuitive, immersive, and effective.

\section{Conclusions}
The proposed algorithm explored in this paper acts as a robust multi-purpose teacher for learning motor tasks, without any parameter adaptation to account for task variation, or the need for extension of the primitive motor gestures. By leveraging its ability to analyze and generate precise patterns of movement, the model can provide real-time feedback, corrective suggestions, and step-by-step guidance to learners engaged in perfecting a motor skill, and accelerate their learning process.

\textbf{Limitations:}
Synthetic Learners: Our validation was primarily based on synthetic learners, which might not entirely emulate the intricacies and unpredictabilities of real human learners. Future studies with human participants are needed to enrich our understanding and provide a more comprehensive picture of the AI teacher's capabilities.

Model Specificity: The AI teacher was specifically trained on the FC task. While it showed adaptability to other motor skills, it might not encompass the vast spectrum of motor skills that exist. More complex or nuanced skills may require specialized training or additional model adjustments.

Environment Design: Our reinforcement learning environment, while comprehensive, might not capture all the nuances of real-world teaching scenarios. Real-world environments can introduce variables and challenges that were not considered in our controlled setting. 

Comparative Analysis: While our AI teacher demonstrated significant potential compared to autonomous learning, we did not compare its effectiveness against seasoned human educators in a side-by-side manner. The AI's performance might differ when juxtaposed directly with such instructors.

Implementation Constraints: The translation of our virtual AI teacher model to physical robots, 
might introduce hardware challenges, and implementation constraints that were not addressed in this study.

\textbf{Future Work:}
This paper presents promising results when tested in a simulated environment with learner agents. The main contribution reported here provides the software apparatus, which is complemented by the effort to develop the physical instrumentation in the context of the EU project that oversees this entire effort\footnote{\url{www.conbots.eu}}. A design of a real-life application of the model, with real-life learners dressed with an exoskeleton, is the next step for assessing the models' practical effectiveness in enhancing fine motor skill acquisition and will provide valuable insights for further refinement and application.
Recent advances with generative AI, have opened new opportunities for movement data generation techniques. 
Generating such data and training an enhanced teacher model could expand its teaching range without needing real-life data and its associated complexity. 

\appendix

\section{Appendix}
\subsection{Target movement}
Let us denote \(\phi=(1/30)\cdot s + \alpha\) as a time step constant, where $s$ denoted the current step number of a $250$ long episode, and $\alpha$ marks a constant offset. Unless stated otherwise, our tests state the offset at the constant value of $3$. Thus, the location of the target takes the exact form
\begin{align}
&\Psi_t[0] = 3 \cdot \sin(1.8 \cdot \phi) + \label{FC_loc_update_gen_0}\\
&3.4 \cdot \sin(1.8 \cdot  \phi) + 2.5 \cdot \sin(1.82  \cdot \phi) + 4.3 \cdot \sin(2.34 \cdot \phi)\notag
\end{align}
\begin{align}
&\Psi_t[1] = 3 \cdot \sin(1.1 \cdot \phi) + \label{FC_loc_update_gen_1}\\
&3.2 \cdot \sin(3.6 \cdot \phi) + 3.8 \cdot \sin(2.5 \cdot \phi) + 4.8 \cdot \sin(1.48 \cdot \phi)\notag
\end{align}

\subsection{Experimental Results}
Other connectivity modality: note we tested several learning step sizes for each modality and came to the conclusion that 1500 is a fair amount for each one. \\
\textbf{Main Experiment: }
The evaluation was performed on 50 episodes for the FC and on 26 episodes for the WESL task. Game units as learning sessions with 1500 steps each, across 10 learners, after outlier removal. 

To confirm the difference in results between the connected and not-connected settings, a series of statistical tests were performed as follows; For each of the four hypotheses, a t-test was conducted on the Frechet distance values of the learner when connected to a virtual teacher, with low-high modality, versus the ones in the not-connected settings. The p-values for the FC task across the four hypotheses are described in Tables \ref{tab:FC_hypth_1} and \ref{tab:wesl_hypth_1}.

\subsection{Robustness tests:} 
To assess the robustness of our AI teacher, we carried out extensive tests under various real-world scenarios. We introduced learners with varying skills, learning rates, and attention levels, and had the AI teach them two tasks. Noise was added to simulate occasional learner errors, with types ranging from Gaussian to random and varying degrees of intensity. All 7 game units were tested, representing different session lengths. 
Results are shown in Tables~\ref{tab:FC_robust} and~\ref{tab:wesl_robust}.

\begin{table}[!ht]\centering
\small
\begin{tabular}{ccccc}\toprule
\textbf{Game} &\textbf{Mean} &\textbf{Mean} &\textbf{p-value} \\
\textbf{Unit} &\textbf{Not-connected} &\textbf{Connected} &\\
\midrule
1 & $0.235$ & $0.524$ & $0.016$\\
2 & $0.437$ & $0.794$ & $<0.5\cdot10^{-4}$ \\
3 & $0.491$& $0.814$ & $<0.0005$\\
4 & $0.427$& $0.874$&$<0.1\cdot10^{-5}$ \\
5 & $0.53$& $0.873$& $<0.0005$ \\
6 & $0.662$& $0.906$ &$0.001$ \\
7 & $0.617$& $0.892$ & $0.005$\\
\bottomrule
\end{tabular}
\caption{H1 results on FC task. Columns $2$ and $3$ show the median Frechet distance at each game unit across all repetitions, and column $4$ shows the p-values for pairwise t-tests.}\label{tab:FC_hypth_1}
\end{table}

\begin{table}[!htp]\centering
\setlength{\tabcolsep}{4pt}
\small
\begin{tabular}{ccccc}\toprule
\textbf{Game Unit} &\textbf{Not-connected} &\textbf{Connected} &\textbf{p-value} \\
\midrule
1 & $0.404$& $0.5$& $0.23$\\
2 & $0.51$& $0.69$& $0.0001$\\
3 & $0.487$& $0.716$& $0.0014$\\
4 & $0.537$& $0.754$ & $<0.5\cdot10^{-4}$\\
5 & $0.528$& $0.739$ & $<0.5\cdot10^{-6}$ \\
6 & $0.56$& $0.79$& $<0.5\cdot10^{-5}$\\
\bottomrule
\end{tabular}
\caption{H1 results on WESL task. Columns $2$ and $3$ show the median Frechet distance at each game unit across all repetitions, column $4$ shows the p-values for pairwise t-tests.}\label{tab:wesl_hypth_1}

\setlength{\tabcolsep}{4pt}
\small
\begin{tabular}{ccccc}\toprule
\textbf{Noise} &\textbf{Noise} &\textbf{Game} &\textbf{Distance} & \textbf{Distance} \\
\textbf{Variance} &\textbf{Type} &\textbf{Unit} &\textbf{Not Connected} & \textbf{Connected} \\
\cmidrule{1-5}
4 &normal &4 &0.62 &0.91 \\
4 &normal &5 &0.63 &0.92 \\\cmidrule{1-5}
4 &uniform &4 &0.63 &0.92 \\
4 &uniform &5 &0.62 &0.91 \\\cmidrule{1-5}
6 &normal &4 &0.67 &0.91 \\
6 &normal &5 &0.71 &0.93 \\\cmidrule{1-5}
6 &uniform &4 &0.69 &0.91 \\
6 &uniform &5 &0.72 &0.89 \\\cmidrule{1-5}
8 &normal &4 &0.61 &0.9 \\
8 &normal &5 &0.65 &0.91 \\\cmidrule{1-5}
8 &uniform &4 &0.63 &0.89 \\
8 &uniform &5 &0.69 &0.89 \\\cmidrule{1-5}
10 &normal &4 &0.61 &0.9 \\
10 &normal &5 &0.7 &0.92 \\\cmidrule{1-5}
10 &uniform &4 &0.51 &0.9 \\
10 &uniform &5 &0.7 &0.92 \\\midrule
\bottomrule
\end{tabular}
\caption{Test results of learners with different capabilities on the FC task.}\label{tab:FC_robust}

\setlength{\tabcolsep}{4pt}
\small
\begin{tabular}{ccccc}\toprule
\textbf{Noise} &\textbf{Noise} &\textbf{Game} &\textbf{Distance} & \textbf{Distance} \\
\textbf{Variance} &\textbf{Type} &\textbf{Unit} &\textbf{Not Connected} & \textbf{Connected} \\
\cmidrule{1-5}
4 &normal &4 &0.48 &0.67 \\
4 &normal &5 &0.55 &0.72 \\\cmidrule{1-5}
4 &uniform &4 &0.58 &0.73 \\
4 &uniform &5 &0.58 &0.73 \\\cmidrule{1-5}
6 &normal &4 &0.56 &0.72 \\
6 &normal &5 &0.54 &0.75 \\\cmidrule{1-5}
6 &uniform &4 &0.55 &0.7 \\
6 &uniform &5 &0.56 &0.74 \\\cmidrule{1-5}
8 &normal &4 &0.56 &0.69 \\
8 &normal &5 &0.59 &0.72 \\\cmidrule{1-5}
8 &uniform &4 &0.56 &0.68 \\
8 &uniform &5 &0.58 &0.74 \\\cmidrule{1-5}
10 &normal &4 &0.57 &0.68 \\
10 &normal &5 &0.56 &0.73 \\\cmidrule{1-5}
10 &uniform &4 &0.54 &0.68 \\
10 &uniform &5 &0.6 &0.74 \\\midrule
\bottomrule
\end{tabular}
\caption{Test results of learners with different capabilities on the WESL task.}\label{tab:wesl_robust}
\end{table}

\section{Acknowledgments}
This project has received funding from the European Union's Horizon 2020 research and innovation programme under grant agreement No 871803. Most prominently, we would like to give tribute for the inception of the interaction model, and respectively for the experimental design and the corresponding data collection, to our project partners UCBM, NU, and ICL. 

\bibliography{aaai24}

\end{document}